\documentclass[10pt,twocolumn,letterpaper]{article}

\usepackage[pagenumbers]{iccv} 
\definecolor{iccvblue}{rgb}{0.21,0.49,0.74}
\usepackage[pagebackref,breaklinks,colorlinks,allcolors=iccvblue]{hyperref}

\usepackage{xcolor}
\usepackage{xr}
\usepackage{pifont}

\def\paperID{3318} 
\def\confName{ICCV}
\def\confYear{2025}

\title{Parametric Shadow Control for Portrait Generation \\ in Text-to-Image Diffusion Models}

\author{
Haoming Cai\textsuperscript{1}\textsuperscript{*}, 
Tsung-Wei Huang\textsuperscript{2}, Shiv Gehlot\textsuperscript{2}, \\
Brandon Y. Feng\textsuperscript{3}, Sachin Shah\textsuperscript{1}, 
Guan-Ming Su\textsuperscript{2}, Christopher Metzler\textsuperscript{1} \\
$^1$University of Maryland \quad $^2$Dolby Labs \quad $^3$MIT
}

\begin{document}

\twocolumn[{%
\renewcommand\twocolumn[1][]{#1}%
\maketitle

\begin{center}
    \captionsetup{type=figure}
    \resizebox{\textwidth}{!}{%
        \includegraphics{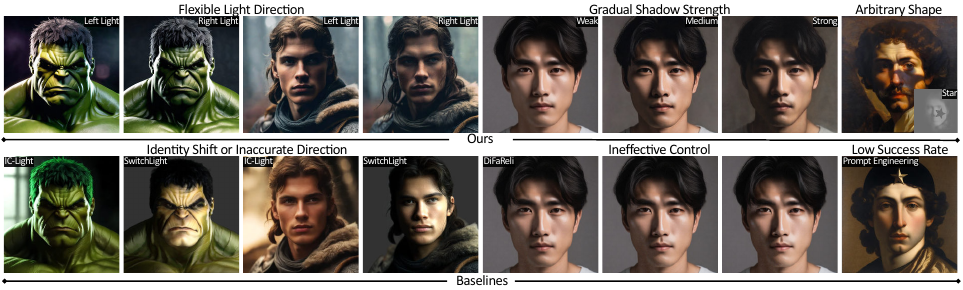}
    }
    \captionsetup{font=small}
    \caption{
     \textbf{Shadow manipulation on generated portrait images with diverse styles challenges existing editing methods.} To address this limitation, we present Shadow Director, which enables intuitive and parametric shadow control during diffusion-based portrait generation, instead of post-processing, across diverse artistic styles. Top: Our method provides parametric control over (left) directional light position, (middle) progressive shadow strength, and (right) arbitrary shadow shapes, while maintaining identity and style consistency. Bottom: Comparison with baseline methods reveals their limitations: SwitchLight \cite{kim2024switchlight} and IC-Light \cite{zhang2025scaling} suffers from identity shifts or inaccurate lighting direction despite requiring substantial training resources, DiFaReli \cite{ponglertnapakorn2023difareli} fails at effectively providing strong variation in shadow strength. Prompt engineering results in unpredictable outcomes.
    }
    \label{fig:teaser}
\end{center}
}]

\begingroup
\renewcommand\thefootnote{*}
\footnotetext{This work was initially conducted during an internship at Dolby Labs.}
\endgroup

\begin{abstract}

Text-to-image diffusion models excel at generating diverse portraits, but lack intuitive shadow control. Existing editing approaches, as post-processing, struggle to offer effective manipulation across diverse styles. Additionally, these methods either rely on expensive real-world light-stage data collection or require extensive computational resources for training.
To address these limitations, we introduce Shadow Director, a method that extracts and manipulates hidden shadow attributes within well-trained diffusion models. Our approach uses a small estimation network that requires only a few thousand synthetic images and hours of training—no costly real-world light-stage data needed.
Shadow Director enables parametric and intuitive control over shadow shape, placement, and intensity during portrait generation while preserving artistic integrity and identity across diverse styles. Despite training only on synthetic data built on real-world identities, it generalizes effectively to generated portraits with diverse styles, making it a more accessible and resource-friendly solution. Homepage: \href{https://www.hm-cai.com/ShadowDirector/}{https://www.hm-cai.com/ShadowDirector/}
\end{abstract}

\section{Introduction}
\label{sec:intro}

Control over visual attributes in AI-generated imagery has advanced significantly \cite{zhang2023adding, luo2024readout}, yet shadow control for portraits remains unexplored in diffusion-based generation.
While current diffusion models excel at generating portraits with consistent lighting based on text prompts, prompt engineering lacks intuitive controls for shadow effects, forcing digital artists to rely on time-consuming trial-and-error prompt engineering.

\subsection{Limitations of Shadow Editing Methods}
\label{subsec:limitations}

Conventional editing-based methods struggle with shadow manipulation in generated images across diverse artistic styles. As shown in Figure \ref{fig:teaser}, these methods not only fail to achieve effective shadow control but also struggle to preserve non-realistic artistic elements such as unique tones in stylized artwork, vibrant color palettes, and subject identity.

This limitation partially arises from restricted diversity and quality in training data. Real-world OLAT (one-light-at-a-time) data by light stage system is prohibitively expensive to collect and inherently limited in diversity. Real-world OLAT datasets primarily capture a small group of real-world subjects, restricting variations in skin tones and facial features while lacking the necessary variety to handle artistic styles. Synthetic data, on the other hand, typically lacks the quality compared to real-world OLAT data.

These limitations in training data have led researchers to incorporate diffusion models with their rich prior information, as seen in methods like DiFaReli \cite{ponglertnapakorn2023difareli} and IC-Light \cite{zhang2025scaling}. DiFaReli attempts to fine-tune diffusion models on synthetic data but fails to unlock the models' full capabilities for diverse artistic styles. IC-Light achieves better performance by scaling up to 10 million training examples (combining real and synthetic data), but struggles with identity preservation and requires substantial computational resources (8 H100 GPUs for a week), making its training pipeline impractical to many researchers. Table \ref{tab:method_tab} summarizes the key aspects and computational resources required for these editing methods.

These challenges motivate us to ask: can we achieve effective shadow manipulation on generated portraits with diverse styles using few synthetic data and computational resources? Recent studies have found that well-trained diffusion models implicitly encode rich information about various visual attributes \cite{du2023generative} like depth, normal \cite{fu2024geowizard}, Shading \cite{kocsis2024lightit}, and human pose \cite{luo2024readout}. This suggests that shadow information for portraits might similarly be embedded within these models, raising an important question: is there an efficient way to access and utilize this shadow-related prior knowledge without extensive retraining?

\subsection{Key Technical Insight and Approach}
\label{subsec:insight}

To address the question of efficiently accessing shadow information in diffusion models, we present Shadow Director: a method for extracting and manipulating shadow information embedded within diffusion models during generation.

Our approach is based on the hypothesis that well-trained diffusion models already encode shadow information within their intermediate latent features. Although this information exists implicitly, we found that a trained shadow estimation network can effectively access it. Specifically, we train a small shadow estimation network that extracts shadow maps from the diffusion model's noisy intermediate features during denoising. Shadow control is implemented through an optimization process during image generation: we calculate the difference between the current estimated shadow (from noisy latent features) and our target shadow, then optimize the latent features to achieve the desired shadow while preserving identity.

Unlike editing methods that work as post-processing, Shadow Director enables parametric and intuitive shadow control directly during the portrait generation process, preserving artistic integrity across diverse styles. Our experiments show that Shadow Director effectively extracts shadow information from diffusion models while requiring only a few thousand synthetic examples of limited quality and just two A6000 GPU with a few hours of training. This resource-efficient design philosophy demonstrates that leveraging information already present in diffusion models can significantly reduce dependence on expensive light-stage data collection, making it feasible for researchers with limited data and computing resources to develop better shadow manipulation methods.

In summary, our contributions include:
\begin{itemize} 
    \item We propose Shadow Director, a novel approach that accesses and manipulates implicit shadow information embedded in the latent space during diffusion model denoising, fundamentally different from editing methods.
    
    \item Our method achieves effective shadow control using only a small synthetic dataset and a few hours of training, demonstrating remarkable data efficiency without depending on costly light-stage data.
    
    \item Shadow Director enables intuitive and parametric control over multiple shadow attributes while preserving identity. Despite training only on synthetic data built on real-world identities, it generalizes to diverse artistic styles where editing methods fail.
\end{itemize}




\section{Related Work}
\label{sec:relatedwork}

\subsection{Portrait Illumination Editing with Neural Network}  
\label{}  

\begin{table}[t]
    \centering
    \resizebox{\linewidth}{!}{%
    \begin{tabular}{lcccc}
        \toprule
        & \textbf{SwitchLight\cite{kim2024switchlight}} & \textbf{IC-Light\cite{zhang2025scaling}} & \textbf{DiFaReli\cite{ponglertnapakorn2023difareli}} & \textbf{Ours} \\ 
        \midrule
        \textbf{Type} & editing & editing & editing & cond Gen \\ 
        \textbf{Philosophy} & neural rendering & scaled dataset & diffusion model & attribute reveal \\ 
        \textbf{Synthetic Data involved} & \checkmark & \checkmark &\checkmark & \checkmark\\
        \textbf{Real-world OLAT involved} & \checkmark & \checkmark & & \\ 
        \textbf{Diffusion Model} &  & \checkmark & \checkmark & \checkmark \\ 
        \textbf{Data Quantities (k)} & 30+ & 10,000+ & 60 & 5 \\ 
        \textbf{GPU} & 32-A6000 & 8-H100 & 1-V100s & 2-A6000 \\ 
        \textbf{Training Time} & 1 week & 140 hours & 8 days & 8 hours \\ 
        
        \bottomrule
    \end{tabular}
    }
    \caption{Comparison with accessible shadow manipulation methods. Key advantages of our approach: (1) Better adaptation to generated images through conditional generation instead of post-processing, (2) No real-world OLAT data required, and (3) Minimal data quantity needed (only 5k samples). Such resource-efficient design proves that our design based on attribute reveal is promising. It enables researchers with limited data and computing resources to more easily develop better shadow manipulation methods in the future.}
    \label{tab:method_tab}
\end{table}

Portrait relighting has been extensively explored in 2D~\cite{pandey2021total, kim2024switchlight, iclight, ren2024relightful, jin2024neural_gaffer, zeng2024dilightnet, sun2019single, wang2020single, paris2003lightweight, shu2017neural, hou2021towards, hou2022face, zhang2020portrait, wang2023sunstage}, and 3D~\cite{rao2024lite2relight, sun2021nelf, tan2022volux, bi2021deep, cai2024real, wang2023sunstage, zhou2019deep, mei2024holo}, often relying on inverse rendering~\cite{barron2014shape, sengupta2018sfsnet} or light-stage data~\cite{debevec2000acquiring}. Physics-based methods~\cite{pandey2021total, kim2024switchlight} decompose albedo, normal, and shading but fail to capture complex effects like subsurface scattering~\cite{kim2022countering, mashita2011measuring, donner2006spectral}. Diffusion-based relighting approaches~\cite{zeng2024dilightnet, jin2024neural_gaffer, iclight, ren2024relightful, hou2024compose, cha2024text2relight} use HDR priors or synthetic datasets but remain constrained by domain gaps~\cite{yeh2022learning, sun2019single, sun2020light}. Models like DiffRelight~\cite{he2024diffrelight} require per-subject fine-tuning for 3D. While SwitchLight~\cite{kim2024switchlight}, IC-Light~\cite{iclight}, and Relightful Harmonization~\cite{ren2024relightful} mitigate gaps through real-image pretraining, pseudo-labeling, and large-scale augmentation.

\subsection{Diffusion Models for Illumination Manipulation}
With the rapid advancement of diffusion models, a wide range of image and video tasks have seen significant progress \cite{ho2020denoising, ho2020denoising, shi2024motionstone, chen2025posta, chen2024repurposing, wang2024flash, wang2023breathing, jia2024dginstyle}. 
While primarily designed for generative tasks, diffusion models implicitly encode a wealth of structural and semantic information within their intermediate representations\cite{luo2024readout, luo2023diffusion, zhang2023talea, tang2023emergent}, which can be leveraged for downstream applications.
Moreover, numerous studies have shown that diffusion models exhibit strong capabilities in estimating and recovering intrinsic attributes such as depth \cite{ke2024repurposing}, surface normals\cite{fu2024geowizard}, and 3D structure\cite{li2023instant3d}, which are valuable for lighting manipulation. 
Recently, researchers have begun exploring how lighting-related information embedded within diffusion models can be utilized \cite{du2023generative, alzayer2024generative}.
Works such as \cite{kocsis2024lightit, xing2024luminet} have demonstrated the feasibility of training diffusion models for illumination manipulation in both indoor and outdoor scenes.  
These studies collectively suggest that well-trained diffusion models inherently encode shadow information. 
This raises the possibility of accessing and utilizing this hidden shadow information to enable shadow control for generated images across diverse artistic styles.

\begin{figure*}[t]
    \centering
    \includegraphics[width=\linewidth]{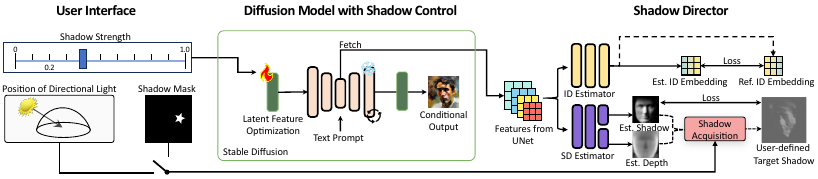}
    \caption{\textbf{Method Overview During Image Generation.} Our approach consists of three main components: 
    (1) \textit{User Interface} (left), which provides intuitive controls for shadow strength, directional light position, and shadow shape;  
    (2) \textit{Diffusion Model with Shadow Control} (middle), where latent features are optimized at selected denoising steps; and  
    (3) \textit{Shadow Director} (right), which extracts shadow and identity information from UNet internal features using two estimators.  
    Shadow Director is trained to infer these attributes from noisy feature maps.  
    During generation, shadow control is achieved through test-time optimization of the latent features (marked with a fire symbol) at early denoising step for larger degree of freedom on shadow manipulation. Before optimization begins, both estimators perform an initial forward pass (dashed lines) to obtain the customized shadow and reference identity embedding. The shadow acquisition process is detailed in Fig.~\ref{fig:method_shadow_acquisition}.
    During optimization, latent features are guided to match the customized shadow while maintaining identity consistency. Notably, the only optimizable component in the pipeline is the latent feature at the selected denoising step. Further architectural details of the estimators and feature extraction are provided in Appendix~\ref{supp_sec:network_architectures}.
}

    \label{fig:method_overview}
\end{figure*}

\begin{figure}[t]
    \centering
    \includegraphics[width=\linewidth]{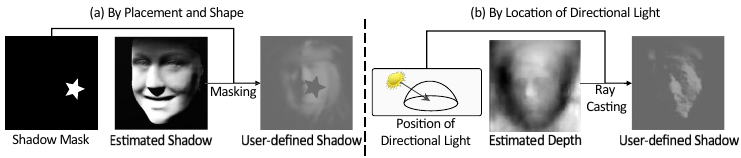}
    \caption{\textbf{Shadow Acquisition.} Two options for customizing shadow maps, which occurs once before latent optimization begins. \textbf{(a)} Shadow placement and shape: A user-defined binary mask is applied to the estimated shadow map. Masked regions become darker in the customized shadow, explicitly defining shadow areas. \textbf{(b)} Shadow synthesis through directional lighting: Using the estimated depth map and user-specified directional light position, we implement ray casting to generate geometrically consistent shadows. Users select only one of these two methods. Detail of ray casting is presented in Appendix.~\ref{supp_subsec:shadow_acquisition}}
    \label{fig:method_shadow_acquisition}
\end{figure}

\begin{figure}[t]
    \centering
    \includegraphics[width=\linewidth]{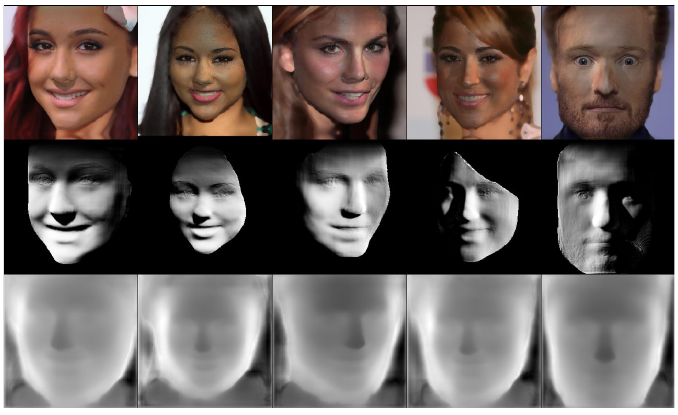}
    \caption{\textbf{Synthetic Training Dataset.} Samples from our synthetic dataset, consisting of paired relit images (top row), shadow maps (middle row), and depth maps (bottom row). Unlike IC-Light requiring 10 million samples, our approach needs only a few thousand paired synthetic examples. These are generated using GeomConsistentFR \cite{hou2022face}. Despite limited photorealism in shadow, this dataset proves sufficient for accessing shadow information embedded within diffusion models. This demonstrates a promising research direction: reducing dependency on expensive light-stage data while focusing on revealing lighting information already hidden in diffusion models.
    }
    \label{fig:method_dataset}
    \vspace{-10pt}
\end{figure}

\section{Method}
\label{method:method}

The key challenge in our approach is accessing and manipulating shadow information embedded within diffusion models during the denoising process, while preserving identity and artistic style.
Our solution, Shadow Director, extracts shadow information from Unet's features and enables intuitive user control through test-time optimization on diffusion model's latent feature maps.
What distinguishes our approach is the ability to reveal and manipulate implicit shadow information during generation with minimal training resources.
Sec.~\ref{method:shadow_director} details the design of Shadow Director for accessing latent shadow information. Sec.~\ref{method:dataset} explains our synthetic training dataset.


\subsection{Shadow Director}
\label{method:shadow_director}

\paragraph{Module Overview.}
\label{method:shadow_director_overview}
As shown in Figure~\ref{fig:method_overview}, Shadow Director comprises two key components: a Shadow-Depth (SD) Estimator and an Identity (ID) Estimator. Both operate on the same UNet's feature maps during denoising. The SD Estimator extracts 2D shadow and depth maps to enable shadow manipulation, while the ID Estimator produces feature embeddings (ID embeddings) to enable identity preservation. Both estimators are implemented as compact neural networks, with architectures details in Appendix~\ref{supp_sec:network_architectures}.

\paragraph{Training of Shadow Director.}
Both estimators are trained independently on our synthetic dataset.
For the SD Estimator, at each training iteration, we generate a noisy latent feature by adding random noise to the clean latent feature of a relit image.
This noisy latent feature is fed into the UNet, and internal features are extracted for shadow and depth estimation.
The SD Estimator is trained to minimize $\mathcal{L}_{\text{SD}} = \mathcal{L}_{\text{L1}}(S_{\text{pred}}, S_{\text{gt}}) + \mathcal{L}_{\text{L1}}(D_{\text{pred}}, D_{\text{gt}})$, where $S$ and $D$ represent shadow and depth maps respectively. For the ID Estimator, each training iteration requires three images: an original image with its original lighting, a same-identity image with different lighting, and a different-identity image.
Similar to SD-Estimator training, we generate noisy latent features by adding random noise to the clean latent features of these three images. Their corresponding ID embeddings are then extracted independently by the ID-Estimator.
We apply a triplet loss \cite{chechik2010large, fu2023dreamsim, luo2024readout} to enforce that embeddings of the same identity become closer while embeddings of different identities are pushed apart. This training approach enables the ID Estimator to maintain consistent identity representations despite lighting variations, ensuring our Shadow Director preserves the subject's identity. We train this ID-Estimator on our synthetic dataset.

\paragraph{Enable Shadow Control.}
Shadow control is achieved through test-time optimization at specific denoising steps.
At a selected early denoising stage, we pause the denoising process and optimize the latent features (UNet input marked with a fire symbol in Figure~\ref{fig:method_overview}) to adjust their hidden shadow attributes.
Before this optimization begins, we obtain the target references (identity and shadow) exactly once (shown red dashed lines in Figure~\ref{fig:method_overview}).
For identity preservation, we extract reference ID embeddings representing the original identity generated by the user-defined text prompt.
For shadow manipulation, we obtain the user-defined target shadow map through either user-defined shadow masks or ray casting, as detailed in Figure~\ref{fig:method_shadow_acquisition}.
Users can control three shadow aspects: Shadow intensity, Shadow shape \& placement, and position of directional light.
Shadow intensity is controlled by the optimization strength parameter.
Shadow placement and shape can be customized through a binary mask (Figure~\ref{fig:method_shadow_acquisition}a).
Alternatively, users can specify a 3D light source position, with shadows generated through ray casting (Figure~\ref{fig:method_shadow_acquisition}b).
The ray casting process is straightforward: for each point on the estimated depth map, a shadow ray is cast toward the light source; if this ray intersects another part of the 3D structure, the point is marked as shadowed.
The test-time optimization employs the following loss function:
$\mathcal{L}_{\text{total}} = \lambda_{\text{shadow}} \cdot \mathcal{L}_{\text{L1}}(S_{\text{current}}, S_{\text{target}}) + \lambda_{\text{identity}} \cdot \mathcal{L}_{\text{sim}}(I_{\text{current}}, I_{\text{ref}})$.
$S_{\text{current}}$ is the current estimated shadow map, $S_{\text{target}}$ is the user-defined target shadow, $I_{\text{current}}$ is the current identity embedding, and $I_{\text{ref}}$ is the reference identity embedding.
The $\mathcal{L}_{\text{sim}}$ represents cosine similarity loss ensuring identity consistency.
The weighting factors $\lambda_{\text{shadow}}$ and $\lambda_{\text{identity}}$ balance shadow control against identity preservation.
Only the latent features are optimized; all network parameters remain frozen.

\begin{figure}[t]
    \centering
    \includegraphics[width=\linewidth]{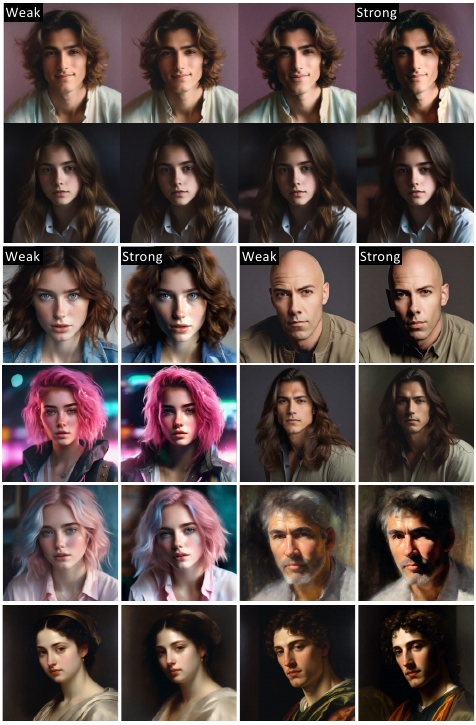}
    \caption{\textbf{Shadow intensity control on generated portraits across diverse styles.}  
The top two rows demonstrate gradual shadow intensity control, while the bottom four rows highlight strong and weak shadow variations for better visual comparison. Shadow Director enables parametric control over shadow strength, ranging from weak to strong, while preserving both identity and artistic integrity.}

    \label{fig:exp_shadow_identity}
\end{figure}

\begin{figure}[h]
    \centering
    \includegraphics[width=\linewidth]{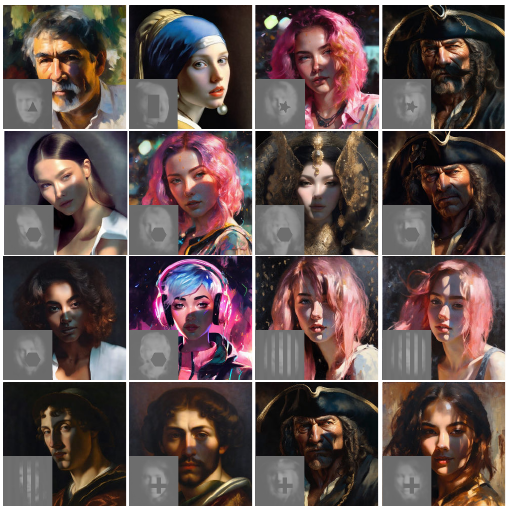}
    \caption{\textbf{Shadow shape control with user-defined masks on diverse portrait images.} Shadow Director enables precise control over shadow shapes and placement using user-defined masks (shown as gray overlays).}
    \label{fig:exp_shadow_masking}
\end{figure}


\subsection{Dataset}
\label{method:dataset}

As shown in Figure~\ref{fig:method_dataset}, our synthetic dataset makes Shadow Director a practical and accessible solution for broader researchers. We construct this dataset using the CelebA dataset~\cite{liu2018large}, generating six lighting variations per identity using GeomConsistentFR~\cite{hou2022face}. This design enables training our ID-Estimator to focus solely on identity features while disregarding lighting variations. Since GeomConsistentFR itself is trained on synthetic data, the generated relit images exhibit limited photorealism. However, our experiments show that this dataset is sufficient for Shadow Director to extract and utilize the rich shadow information already encoded in diffusion models. In this work, we focus on maximizing the potential of limited but accessible synthetic data in conjunction with a well-trained diffusion model.

\begin{figure*}[t]
    \centering
    \includegraphics[width=\textwidth]{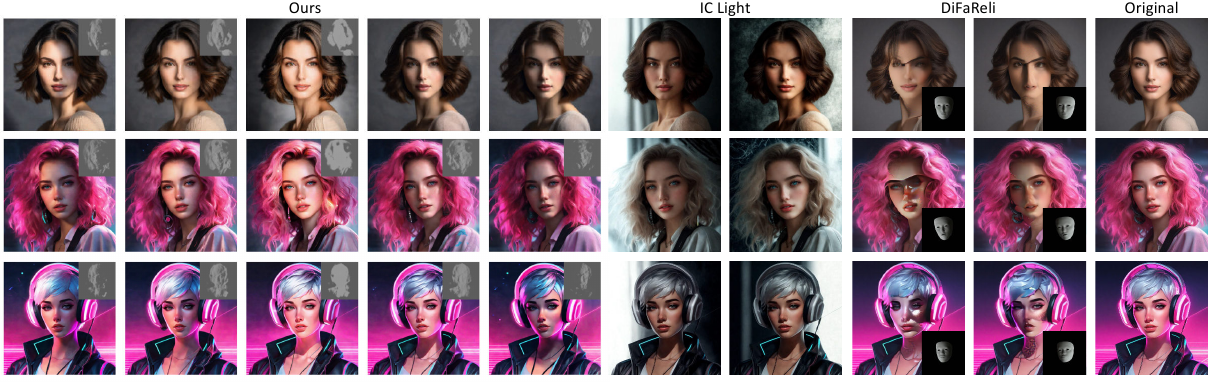}
    \caption{
     \textbf{Shadow Synthesis by Controlling Lighting Direction: Comparison with Diffusion-Based Methods.} Our approach (leftmost 5 columns) consistently preserves identity and artistic style while providing varied shadow control based on specified lighting positions. In contrast, diffusion-based editing methods IC-Light (columns 6-7) and DiFaReli (columns 8-9) show significant identity shifts and style inconsistencies compared to the original portraits (rightmost column). Results are shown across three different portrait styles: realistic (top row), super-realistic (middle row), and cartoon (bottom row). Our method maintains vibrant colors, facial features, and artistic elements while providing intuitive shadow control by directional lighting position. Please see Appendix \ref{supp_sec:additional_results} for more lighting positions.
    }
    \label{fig:exp_shadow_synthesis}
\end{figure*}

\begin{figure*}[h]
    \centering
    \includegraphics[width=\textwidth]{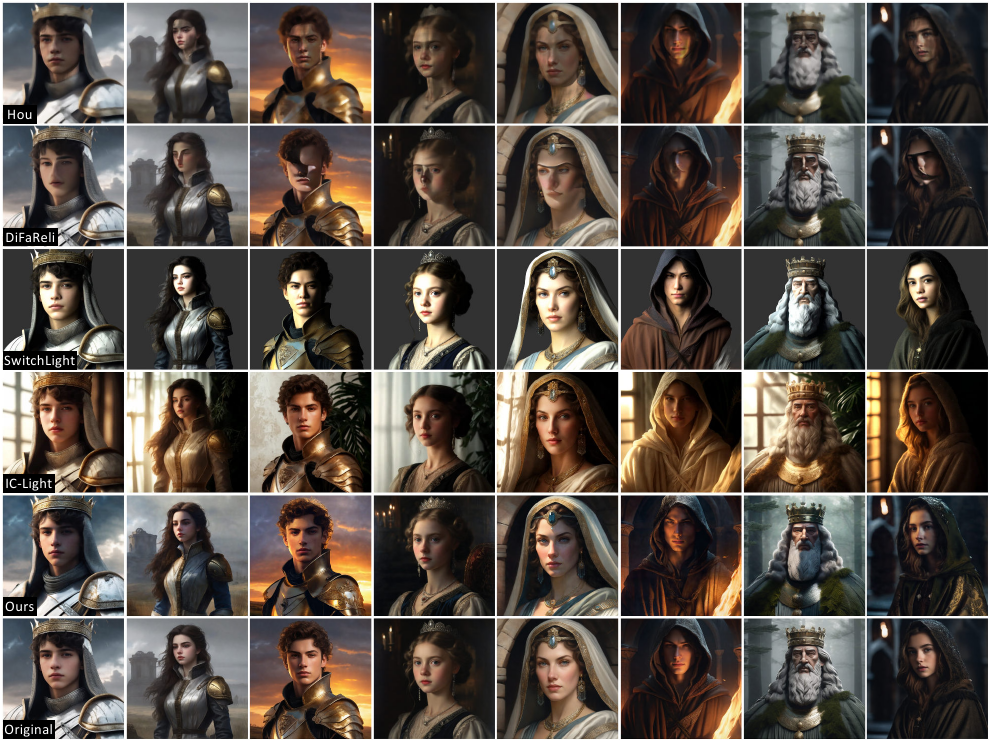}
    \caption{\textbf{Shadow Synthesis by Controlling Lighting Direction.} Comparison of portrait relighting across different editing methods generated portrait. Each row corresponds to a different method, while each column maintains left lighting direction for convenient visual comparison. For SwitchLight, we create a related environment map to simulate directional lighting.}
    \label{fig:exp_shadow_synthesis_02}
\end{figure*}

\section{Experiment}
Additional qualitative results demonstrating Shadow Director's control are provided in Appendix \ref{supp_sec:additional_results}. Detailed implementation is provided in Appendix \ref{supp_sec:implementation_details}.

\subsection{Settings of Training and Inference}
\label{exp:settings}

We implement our pipeline based on Stable Diffusion XL (SDXL) 1.0. 
For the synthetic dataset, we randomly select 1,000 identities from the CelebA dataset, rendering each with 6 different lighting conditions using GeomConstantFR. This generates a total of 6,000 synthetic images, which we split into 5,700 for training and 300 for validation to evaluate the SD Estimator's prediction accuracy.
Both the SD Estimator and ID Estimator are trained on a single NVIDIA A6000 GPU, with each estimator taking approximately 8 hours to train. We use the Adam optimizer with a learning rate of 1e-4 and a batch size of 8.
For inference-time shadow control, we begin latent feature optimization at denoising step 40 out of 100 (corresponding to t=0.6). We use the Adam optimizer with a learning rate of 2e-4 for latent feature optimization. Our method adds only about 20 percent additional computation time compared to standard SDXL inference.

\subsection{Shadow Strength}

Our approach enables users to gradually adjust shadow intensity while preserving surrounding lighting conditions. Unlike text-prompt engineering, which lacks precise control over intensity changes, our provides fine-intuitive control over shadow strength. As shown in Fig.\ref{fig:exp_shadow_identity}, our method achieves modulation of shadow strength through latent optimization during inference time. Shadow intensity correlates directly with the number of optimization iterations applied, allowing for gradual adjustments based on user preference.

\subsection{Shadow Placement and Shape}
Our method offers intuitive control over shadow placement and shape through user-defined shadow mask. In contrast to existing methods that generate unrealistic blended synthetic shadows, our approach ensures natural shadow placement and integration. Results in Fig.\ref{fig:exp_shadow_masking} demonstrate controlled shadow placement using custom shadow masks.

\subsection{Position of Directional Light}
\label{exp:directional_light}

Our method allows users to specify the 3D position of the directional light source for portrait relighting. As shown in Fig.~\ref{fig:exp_shadow_synthesis}, Shadow Director generates geometrically consistent shadows while preserving identity and artistic style. In contrast, other diffusion-based methods such as IC-Light and DiFaReli struggle with effective shadow control and often introduce identity shifts.

To demonstrate our method's performance across a broad range of portraits, we randomly generate 800 portraits using text prompts (detailed in Appendix) without specifying lighting conditions.  
For each portrait, all methods generate two relit images under left and right directional lighting correspondingly. 

For qualitative evaluation, we present results in Fig.~\ref{fig:exp_shadow_synthesis_02}. 
Comparisons are made against IC-Light~\cite{zhang2025scaling}, SwitchLight~\cite{kim2024switchlight}, Hou et al.~\cite{hou2022face}, and DiFaReli~\cite{ponglertnapakorn2023difareli}.
For quantitative evaluation, we partially follow the protocols in \cite{cha2024text2relight} and \cite{ren2024relightful}, assessing both objective metrics (Table~\ref{tab:quantitative_main}) and user studies (Table~\ref{tab:user_study}).
Since ground-truth relit images are unavailable, non-reference image quality assessment methods are used to measure realism and image fidelity. CLIP-IQA+\cite{ wang2023exploring}, NIMA\cite{talebi2018nima}, and ARNIQA \cite{agnolucci2024arniqa} assess image authenticity to ensure that shadow manipulation does not degrade perceptual quality. Following \cite{cha2024text2relight}, CLIP Vision Score (CVS) is used to measure identity preservation through cosine similarity between the clip embedding \cite{radford2021learning} of source and relit images. CLIP Vision-Text Alignment (CVTA) evaluates how well relit images align with the augmented text prompts specifying "left light" or "right light." 
To further validate the results, we conduct user studies (Table~\ref{tab:user_study}) to assess shadow consistency and identity preservation through subjective evaluation. 43 users and 20 samples involved. For each sample, order of methods is random.

\subsection{Ablation Study}
We further investigate the necessity of each component in Shadow Director through a series of ablation studies. These studies are summarized in Table~\ref{tab:exp_ablation_class}, with corresponding quantitative results presented in Table~\ref{tab:exp_ablation_quantitative}.

\vspace{1\baselineskip} 

\noindent\textbf{Optimal Denoising Step for Shadow Control.}
We investigate whether shadow control is most effective during early/middle denoising steps, later steps, or only at the final step. Our experiments confirm that applying shadow control during early and middle denoising steps (t=0.5-0.7) provides the optimal balance between shadow manipulation flexibility and image quality preservation, outperforming later-stage or final-step interventions.

\vspace{1\baselineskip} 

\noindent\textbf{Input of Shadow Estimator.}
We examine which feature source provides the most effective shadow information by comparing three options: UNet input, UNet internal features, and UNet output. Our experiments confirm that UNet internal features yield better performance due to their richer information. Consistent with findings in recent studies \cite{luo2024readout}

\begin{table}[t]
    \centering
    \caption{Quantitative results on 800 unseen portrait images generated from text prompts. 'Ori' refers to the original image, serving as the upper bound. Our method is highlighted in bold.}
    \vspace{-10pt}
    \resizebox{1\columnwidth}{!}{%
    \begin{tabular}{lcccccc}
        \toprule
        \textbf{Metric} & \textbf{Ori} & \textbf{Hou 2022}~\cite{hou2022face} & \textbf{DiFaReLi}~\cite{ponglertnapakorn2023difareli} & \textbf{IC-Light}~\cite{zhang2025scaling} & \textbf{SwitchLight}~\cite{kim2024switchlight} & \textbf{Ours} \\
        \midrule
        \textbf{CVTA} $\uparrow$ & 0.3307 & 0.2629 & 0.2466 & 0.2606 & 0.2520 & \textbf{0.2692} \\
        \textbf{CVS} $\uparrow$ & 1.0000 & 0.9386 & 0.9049 & 0.9140 & 0.8945 & \textbf{0.9460} \\
        \textbf{CLIPIQA$+$} $\uparrow$ & 0.7366 & 0.3747 & 0.3906 & 0.4185 & 0.3876 & \textbf{0.4484} \\
        \textbf{NIMA} $\uparrow$ & 6.3738 & 5.0195 & 5.2370 & 5.7193 & 5.3219 & \textbf{5.7735} \\
        \textbf{ARNIQA} $\uparrow$ & 0.6895 & 0.4603 & 0.4831 & 0.5236 & 0.4442 & \textbf{0.5480} \\
        \bottomrule
    \end{tabular}%
    \label{tab:quantitative_main}
}
\end{table}

\begin{table}[t]
    \centering
    \vspace{-10pt}
    \caption{User study for preference. 2.44\% selecte 'None'}
    \vspace{-10pt}
    \resizebox{1\columnwidth}{!}{%
    \begin{tabular}{lccccc}
        \toprule
          & Ours & IC-Light & SwitchLight & DiFaReli & Hou \\
        \midrule
        Preference Rate & 45.69\% & 27.44\% & 20.46\% & 1.86\% & 2.09\% \\
        \bottomrule
    \end{tabular}%
    
    \label{tab:user_study}
    }
    \vspace{-15pt}
\end{table}

\vspace{1\baselineskip} 

\noindent\textbf{Latent-space vs. Noisy RGB-space Estimation.}
This study compares our latent-space shadow estimation approach against an alternative RGB-space pipeline. The alternative first generates noisy RGB images from the UNet's output at early denoising steps, then applies an RGB-based shadow estimator (like our data generation method GeomConstantFR) to obtain noisy shadow and depth maps. After acquiring the user-defined shadow through a similar process, the loss between this target and the estimated noisy shadow in RGB space is backpropagated through both the RGB estimator and UNet to optimize the latent feature map. Although conceptually similar to our approach, this RGB-space alternative introduces reliability issues during latent optimization. Our experiments demonstrate that latent-space estimation achieves better shadow quality and identity preservation, confirming our design choice to operate directly on UNet's internal feature maps rather than noisy RGB outputs from denoising step.

\vspace{1\baselineskip} 

\noindent\textbf{Necessity of ID Embedding for ID Preservation.}
We demonstrate why our dedicated ID embedding approach is superior to the simpler alternative of applying L1 loss directly on latent features. In this alternative approach, the SD-Estimator remains, but identity preservation is attempted by constraining the optimized UNet input/output features to remain close to their original values via L1 distance. Unlike this direct feature constraint, our ID embedding network specifically captures meaningful identity characteristics while allowing shadow-relevant features to change. Our experiments confirm that specialized identity embeddings are essential for maintaining consistent identity during shadow manipulation, as direct feature constraints either limit shadow flexibility or fail to preserve key identity elements.

\vspace{1\baselineskip} 

\begin{figure}[t]
    \centering
    \includegraphics[width=\linewidth]{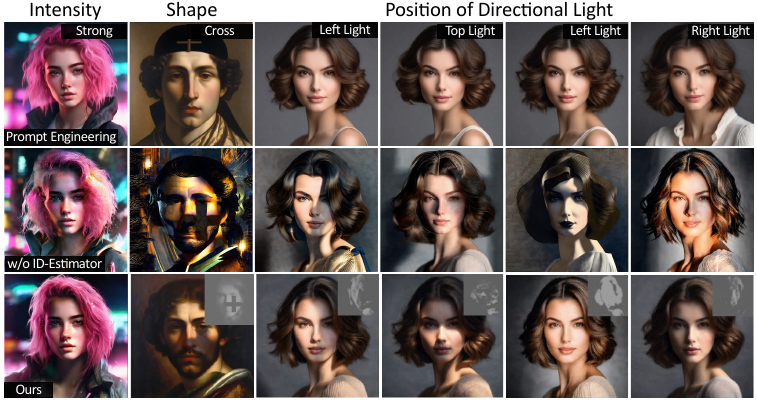}
    \caption{\textbf{Necessity of our ID-Estimator and failure of naive prompt engineering.}  The first row shows that Prompt Engineering fail to produce consistent or controllable shadows. The second row highlights artifacts and identity inconsistencies without our ID-Estimator. The third row (Ours) demonstrates that our full method achieves effective shadow manipulation while preserving identity and style.}
    \label{fig:exp_ablation_id_neccesity}
\end{figure}

\noindent\textbf{Necessity of ID-Estimator.}
Without the ID Estimator to constrain latent optimization, the generated image tends to overemphasize shadow optimization, often resulting in unexpected textures and identity distortions. As shown in Fig.\ref{fig:exp_ablation_id_neccesity}, optimizing for strong shadows without identity preservation loss leads to significant unintended artifacts.


\begin{table}[t]
\caption{
\textbf{Necessity of Design in Shadow Director.}
}
\vspace{-10pt}
\label{tab:exp_ablation_class}
\centering
\resizebox{\columnwidth}{!}{%
\setlength{\tabcolsep}{3.5pt}
\begin{tabular}{@{}lccccccc}
\toprule
                                & \multicolumn{2}{c}{Feature Extraction Location} & \multicolumn{3}{c}{Denoising Step for Latent Optim} & \multicolumn{2}{c}{Constraint Choice} \\ 
                                
                                \cmidrule(l){2-3} \cmidrule(l){4-6} \cmidrule(l){7-8}
                                
 & Unet's Internal & Unet's Output & Early and Middle & Latter & Last One & SD & ID\\ 
\midrule
(a) Ours  & $\checkmark$ & - & $\checkmark$ & - & - & SD-Estimator & ID-Estimator\\

\midrule
\multicolumn{8}{c}{\textbf{Ablation Study 1: Optimal Denoising Step for Shadow Control}} \\
\midrule
(b)  & $\checkmark$ & - & - & $\checkmark$ & - & SD-Estimator & ID-Estimator\\
(c)  & $\checkmark$ & - & - & - & $\checkmark$ & SD-Estimator & ID-Estimator\\
\midrule

\multicolumn{8}{c}{\textbf{Ablation Study 2: Input of Shadow Estimator }} \\
\midrule
(d)  & - & $\checkmark$ & $\checkmark$ & - & - & SD-Estimator & ID-Estimator\\

\midrule
\multicolumn{8}{c}{\textbf{Ablation Study 3: Latent-space vs. Noisy RGB-space Estimation}} \\
\midrule
(e)  & $\checkmark$     & $\checkmark$     & - & - & - & RGB-Estimator & ID-Estimator\\
     &  (for Ours-ID) & (for RGB-SD)   &   &   &   &     &     \\

\midrule
\multicolumn{8}{c}{\textbf{Ablation Study 4: Necessity of ID Embedding for ID Preservation}} \\
\midrule
(f)  & $\checkmark$    & - & - & - & - & SD-Estimator & L1 Loss\\
(g)  & $\checkmark$    & $\checkmark$  & - & - & - & SD-Estimator & L1 Loss\\
     & (for Ours-SD) & (for L1-ID)   &   &   &   &     &     \\

\bottomrule
\end{tabular}%
}
\vspace{-10pt}
\end{table}


\begin{table}[t]
    \centering
    \caption{\textbf{Necessity of Design in Shadow Director.}}
    \vspace{-10pt}
    \label{tab:exp_ablation_quantitative}
    \resizebox{1\columnwidth}{!}{%
    \begin{tabular}{lccccccc}
        \toprule
        \textbf{Metric} & \textbf{(a) Ours} & Method \textbf{(b)} & Method \textbf{(c)} & Method \textbf{(d)} & Method \textbf{(e)} & Method \textbf{(f)} & Method \textbf{(g)} \\
        \midrule
        \textbf{CVTA}      $\uparrow$ & \textbf{0.2692} & 0.2410 & 0.2356 & 0.2197 & 0.1975 & 0.1556 & 0.1485\\
        \textbf{CVS}      $\uparrow$ & \textbf{0.9460} & 0.9325 & 0.9320 & 0.8945 & 0.8173 & 0.7685 & 0.7679\\
        \textbf{CLIPIQA$+$} $\uparrow$ & \textbf{0.4484} & 0.4325 & 0.4392 & 0.3639 & 0.3220 & 0.2873 & 0.2682\\
        \bottomrule
    \end{tabular}%
    }
    \vspace{-10pt}
\end{table}

\begin{figure}[h]
    \centering
    \vspace{-10pt}
    \includegraphics[width=\linewidth]{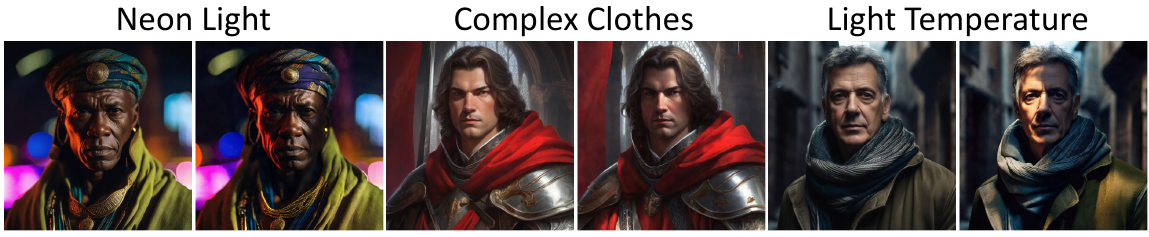}
    \vspace{-14pt}
    \caption{\textbf{Limitations and Future Work.} Each group shows images without (left) and with (right) shadow control. 
    (a) Strong lighting effects in text prompts, such as neon lighting, may override Shadow Director’s control.  
    (b) Training on CelebA, which consists mostly of simple, straight-on portraits, makes it difficult to preserve intricate clothing details in more complex outfits.  
    (c) The model does not explicitly handle lighting temperature, which may lead to color tone inconsistencies.}
    \vspace{-12pt}
    \label{fig:exp_shadow_masking}
\end{figure}

\section{Discussion}
\label{sec:discussios}
\paragraph{Limitation.} Shadow Director is trained on CelebA, which mainly consists of close-up, straight-on portraits with simple clothing, making intricate garment preservation in full-body images challenging. Expanding dataset diversity or constraining latent optimization could help mitigate this. Additionally, strong lighting effects in text prompts, such as neon lighting, may override Shadow Director’s control, which could be improved through refined hyperparameter tuning. The model also lacks explicit lighting temperature control, sometimes causing color tone inconsistencies. Despite these challenges, Shadow Director demonstrates that effective shadow control can be achieved with accessible training resources, paving the way for improvements with broader datasets and enhanced lighting models.

\paragraph{Relation to Shadow Removal.}  
Shadow Director does not perform shadow removal on the original image, as it does not reconstruct occluded details. Instead, it is a conditional generation framework that guides the diffusion model to generate or suppress shadows in specific region during synthesis. Unlike shadow removal methods that recover lost information, Shadow Director conditions the model to produce lighting-consistent images. For example, in the eighth column of Fig.~\ref{fig:exp_shadow_synthesis_02}, the left cheek shadow disappears not due to removal but because the model, conditioned on new lighting, generates an image where the shadow no longer forms. This distinction highlights Shadow Director’s role in controlling shadow generation rather than removing it.

\section{Conclusion}  
We introduced Shadow Director, a diffusion-based approach for intuitive shadow control in portrait generation. By leveraging latent shadow information, our method enables parametric control while preserving identity across diverse styles. Shadow Director achieves this with minimal training data and computation, demonstrating a resource-efficient way for shadow control in diffusion models.

\noindent\textbf{Acknowledgment.} H.C. and C.A.M. were supported in part by gift funds from Dolby and a UMD Grand Challenges Seed Grant.

{
    \small
    \bibliographystyle{ieeenat_fullname}

}

\clearpage
\setcounter{page}{1}
\maketitlesupplementary

\setcounter{section}{0}
\renewcommand{\thesection}{S\arabic{section}}

\section{Overview}
In this Appendix, we present:
\begin{itemize}
    \item Section \textcolor{red}{2}: Network architectures.
    \item Section \textcolor{red}{3}: Implementation details.
    \item Section \textcolor{red}{4}: Additional Results.
    \item Section \textcolor{red}{5}: Evaluation Details.
    \item Section \textcolor{red}{6}: Detailed Text Prompt
\end{itemize}

\section{Network Architectures} \label{supp_sec:network_architectures}

In Fig.\ref{supp_fig:Fig_supp_Details}, we present the detailed network architecture of our three key components: the intermediate feature extraction process, the shadow-depth estimator, and the identity estimator. While shadow-depth estimator and the identity estimator are trained separately, they work in conjunction during the inference phase's latent optimization process.

Our design draws inspiration from Readout\cite{luo2024readout} and Diffusion Hyperfeatures\cite{luo2023diffusion}. While these works demonstrated that latent optimization through a compact network can effectively control attributes like human pose and placement, we demonstrate that this control philosophy can be effectively extended to manipulate intrinsic properties like shadows.

\subsection{Detail of Intermediate Feature Fetch} \label{supp_subsec:intermediate_feature_fetch}
The leftmost panel of Fig.\ref{supp_fig:Fig_supp_Details} illustrates our feature extraction process from UNet. We fetch intermediate features at multiple scale in the Unet to capture rich shadow-related information embedded in the model.

\subsection{Detailed Architecture of SD Estimator} \label{supp_subsec:sd_estimator}
The middle and right panels in Fig.\ref{supp_fig:Fig_supp_Details} show our shadow-depth estimator architecture. It begins with feature fusion through a series of convolutional layers, followed by weighted sum aggregation. The output layers use multiple convolutions to predict shadow and depth maps.

\subsection{Detailed Architecture of ID Estimator} \label{supp_subsec:id_estimator}
\begin{figure}[t]
    \centering
    \includegraphics[width=\linewidth]{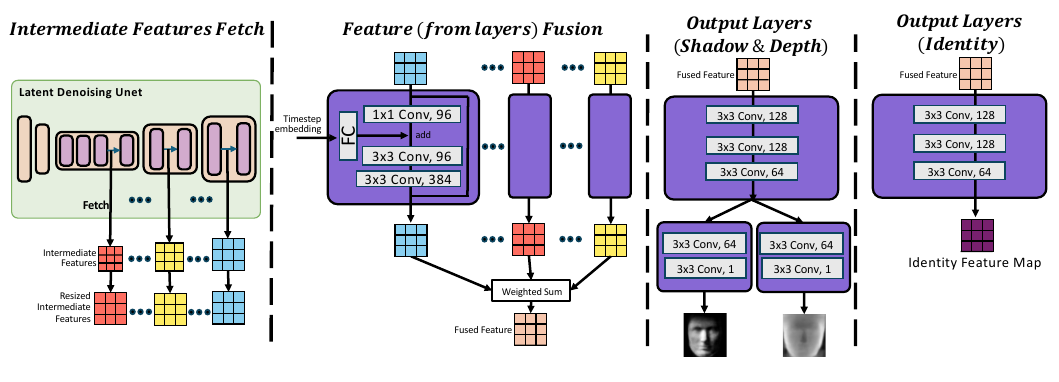}
    \caption{\textbf{Details of network architecture.} Both shadow-depth estimator and identity estimator are trained separately but work jointly during inference-time optimization. The pipeline consists of three main steps: (1) Intermediate Features Fetch: extracts multi-scale features from Latent Denoising UNet timesteps as shared input. (2) Feature Fusion: processes features through convolutional layers with weighted sum aggregation. (3) Attributes Output : For Shadow \& Depth Output Layers, it generates explicit shadow and depth maps through convolutional layers. For Identity Output Layers, it uses same feature processing structure but outputs identity feature map directly for loss computation.}
    \label{supp_fig:Fig_supp_Details}
\end{figure}

The rightmost panel in Fig.\ref{supp_fig:Fig_supp_Details} depicts our identity estimator, which shares a similar convolutional structure but is specifically designed to extract and maintain identity-related features. Though trained separately from the shadow-depth estimator, it processes the same input features. Unlike the shadow-depth estimator that outputs explicit attribute maps, this network directly produces an identity feature map that guides portrait characteristic preservation during optimization.

\section{Implementation Details} \label{supp_sec:implementation_details}

\subsection{Shadow Acquisition Details} \label{supp_subsec:shadow_acquisition}
\begin{figure}[t]
    \centering
    \includegraphics[width=\linewidth]{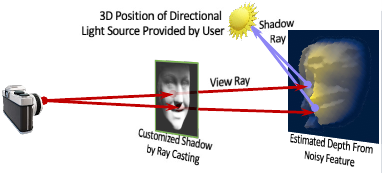}
    \caption{Customized shadow acquisition using ray casting. The user specifies the 3D position of the light source, and the shadow is generated based on a depth map estimated by the Shadow Director during inference.}
    \label{fig:shadow_acquisition_raycast}
\end{figure}

For the ray-casting option, we use the shadow generation algorithm from \cite{hou2022face}. This algorithm takes a depth map and a user-defined lighting position in 3D space as inputs. In brief, a shadow ray is cast toward the light source for each point on the estimated depth map (Figure \ref{fig:shadow_acquisition_raycast}). This depth map is generated by the Shadow Director during inference. If the ray intersects another part of the 3D structure, the point is marked as shadowed. This method ensures that shadows align accurately with the 3D geometry. 

\subsection{Training Phase Details} \label{supp_subsec:training_phase}
Our framework is built on the stable-diffusion-xl-base-1.0 model. In training phase, the maximum time step is 1000. In each training iteration, we randomly add noise to clean latent features through the scheduler. Shadow-depth estimator and identity estimator are trained separately but share identical training settings: learning rate of $1 \times 10^{-3}$, zero weight decay. For the shadow-depth estimator, we use L1 loss to supervise both depth and shadow map predictions against their ground truth (synthetic dataset). And the batch size is 8. For the identity estimator, we employ a hinge loss with 0.5 margin to ensure positive samples remain close in feature space. Moreover, the batch size is 3. 2 positive samples and 1 negative sample. Figure \ref{} straightforwardly illustrates the training mechanism of ID-Estimator.

\subsection{Inference Phase Details} \label{supp_subsec:inference_phase}
Our shadow manipulation pipeline provides three user control parameters: input type (either binar y mask or 3D light position), and shadow strength (ranging from 0 to 1). A shadow strength of 0 maintains current shadows, while 1 triggers maximum manipulation with 30 optimization iterations. To reduce shadow strength, users can add light to desired regions using 3D light positioning.

The manipulation process involves one generation round with 100 time steps. Shadow manipulation occurs specifically at time step 40. At this step, we first generate the depth map and create a customized shadow map based on user's setting. We then optimize the unconditional branch latent features using Adam optimizer with learning rate 5e-2. The number of optimization iterations is determined by the user-specified shadow strength (strength × 30). During optimization, we combine identity and shadow losses with weights of 3 and 1 respectively. The CFG scale is set to 6 throughout the process.

\section{Additional Experimental Results} \label{supp_sec:additional_results}

\begin{figure*}[!htbp]
    \centering
    \includegraphics[width=\linewidth]{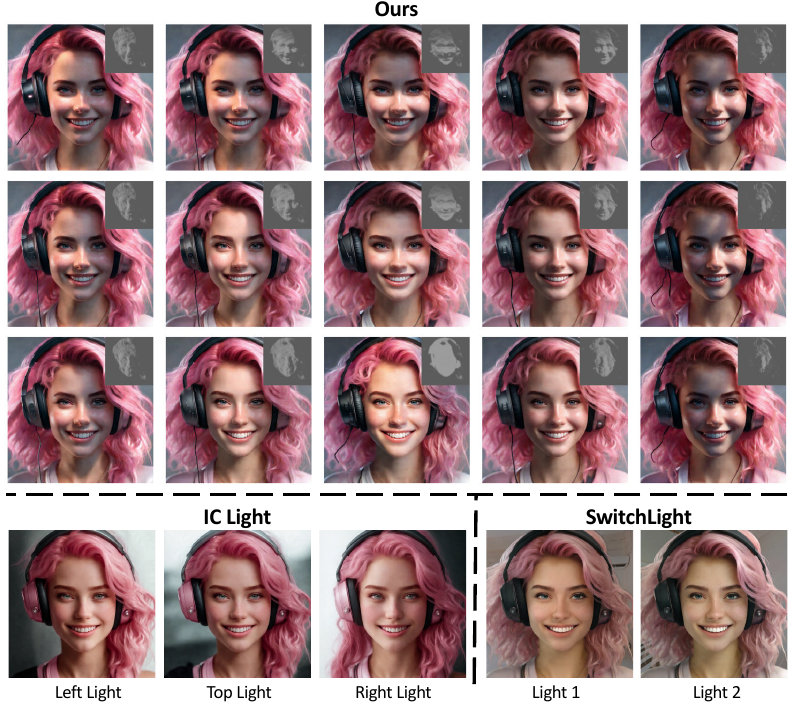}
    \caption{Shadow Synthesis under more lighting conditions. SwitchLight does not use directional lighting here, highlighting its limitation in identity preservation.}
    \label{supp_fig:re_lighting_figure_supp_07}
\end{figure*}

\begin{figure*}[!htbp]
    \centering
    \includegraphics[width=\linewidth]{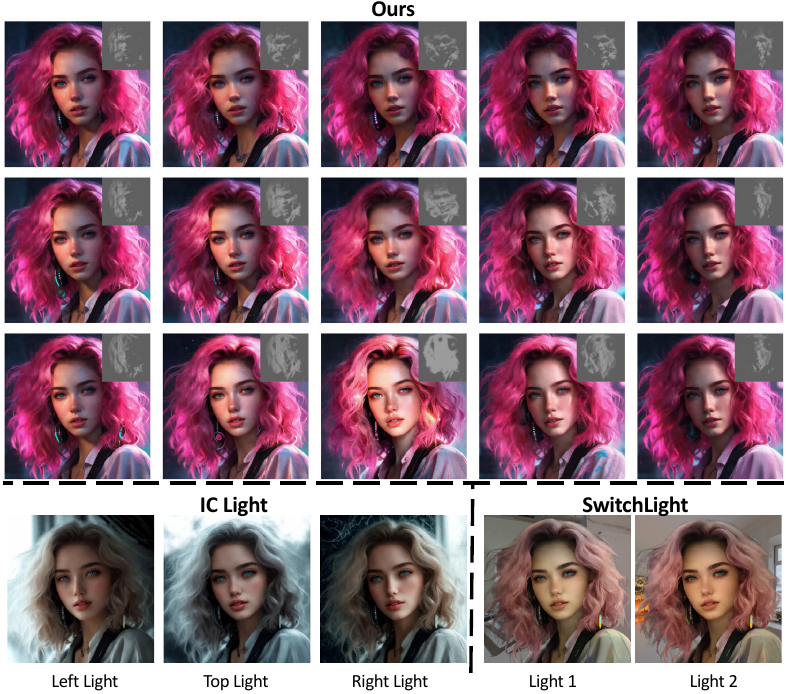}
    \caption{Shadow Synthesis under more lighting conditions. SwitchLight does not use directional lighting here, highlighting its limitation in identity preservation.}
    \label{supp_fig:re_lighting_figure_supp_08}
\end{figure*}

\begin{figure*}[!htbp]
    \centering
    \includegraphics[width=\linewidth]{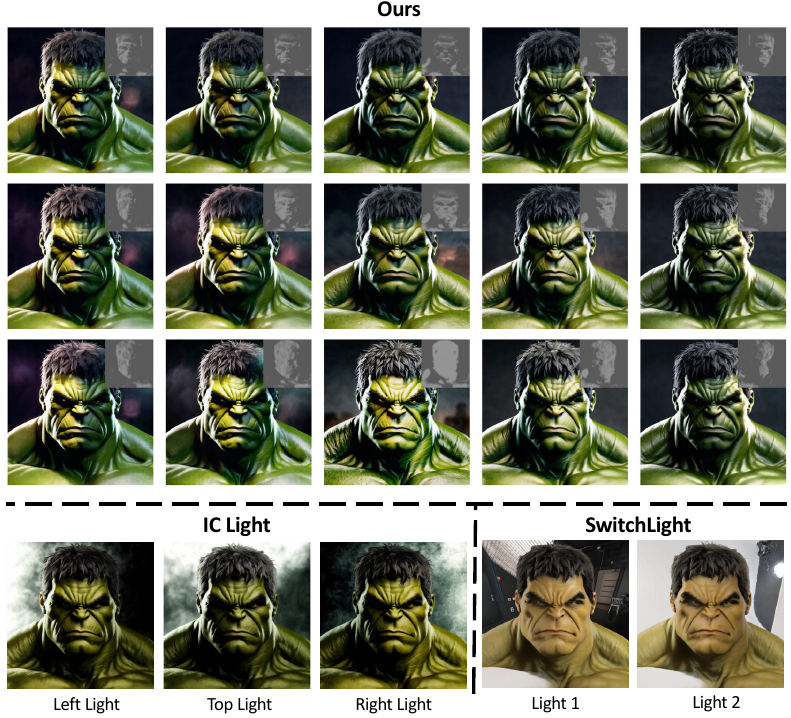}
    \caption{Shadow Synthesis under more lighting conditions.SwitchLight does not use directional lighting here, highlighting its limitation in identity preservation. For IC-Light here, "Hulk" text prompt is feed into as another condition to help IC-Light maintain the identity}
    \label{supp_fig:re_lighting_figure_supp_09}
\end{figure*}

\begin{figure*}[!htbp]
    \centering
    \includegraphics[width=\linewidth]{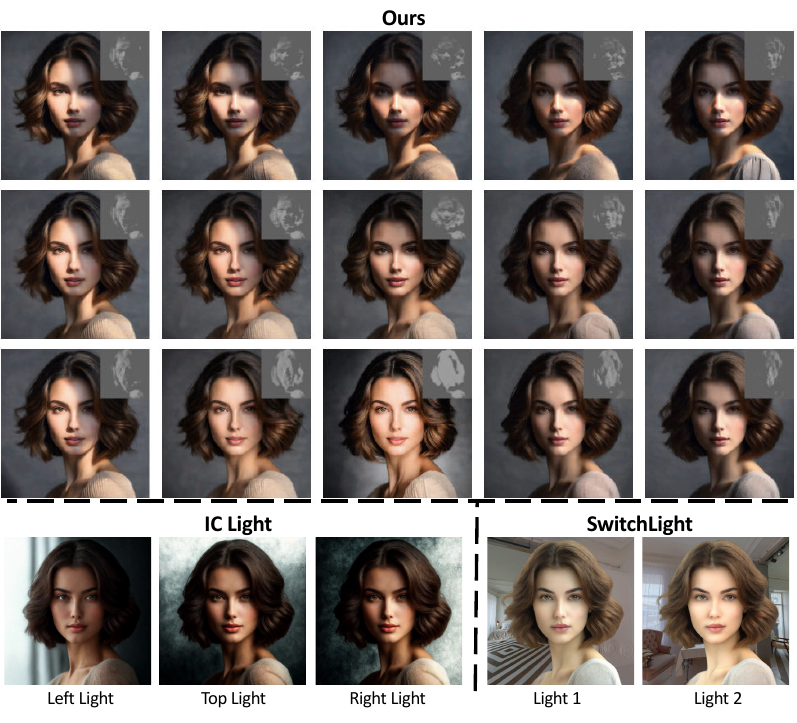}
    \caption{Shadow Synthesis under more lighting conditions. SwitchLight does not use directional lighting here, highlighting its limitation in identity preservation.}
    \label{supp_fig:re_lighting_figure_supp_10}
\end{figure*}

\begin{figure*}[!htbp]
    \centering
    \includegraphics[width=\linewidth]{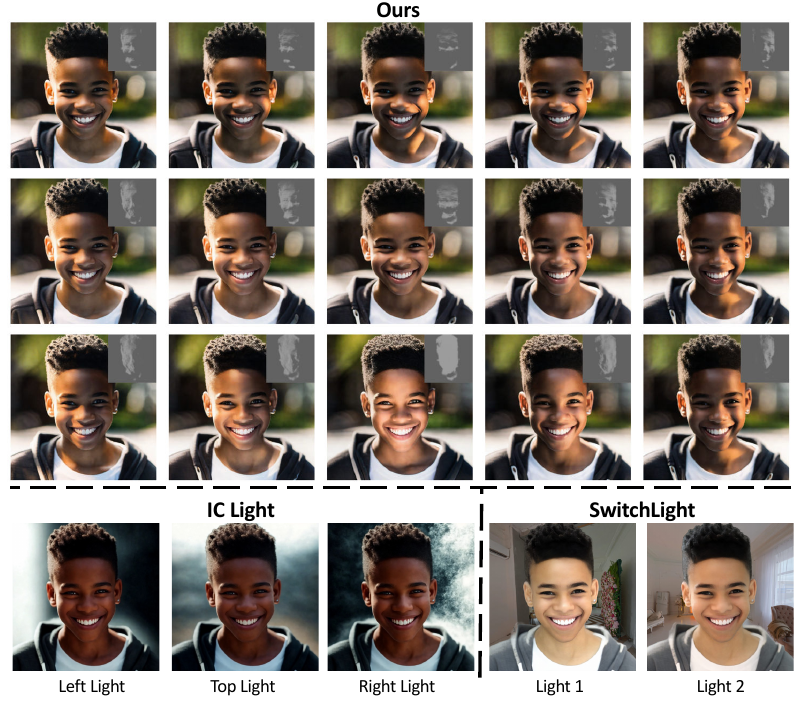}
    \caption{Shadow Synthesis under more lighting conditions. SwitchLight does not use directional lighting here, highlighting its limitation in identity preservation.}
    \label{supp_fig:re_lighting_figure_supp_11}
\end{figure*}

We provide additional results on Shadow Synthesis via Lighting Position Control in Figures \ref{supp_fig:re_lighting_figure_supp_01}–\ref{supp_fig:re_lighting_figure_supp_11}.

\section{Evaluation Details} \label{supp_sec:evaluation_details}
\subsection{Baseline Implementation Details} 
In Hou \cite{hou2022face}
 method, one may find there is distortion around face. We found that paper, like DiFaReLi, when they reimplement Hou's method. Also has similar distortion. Therefore, it comfirm our implemention correctness. For DiFaReli, after installation, we test author's demo to obtain visually exact same result. However, we found that DiFaReli cannot push the shadow strength be harder. The range of enable shadow strength is relatively limited, indicating this is a challenging task. Meanwhile, we find that the all those required element as needed input for DiFaReli looks reasonable, comfirming our re-implementation is correct. In specific, to do relighting with DiFaReli, there should be a source img to provide refernece shadow. We found that this target shadow is correctly transfer to the relit image's related input, as we shown in the main paper experiment figure. Notably, we use very simple portrait and official demo portrait image as target shadow. However, the DiFaReLi still fails on generated image, indicating this taks is a challenge one.

\subsection{User Study} \label{supp_subsec:user_study}
\begin{figure}[tbh!]
    \centering
    \includegraphics[width=0.7\linewidth]{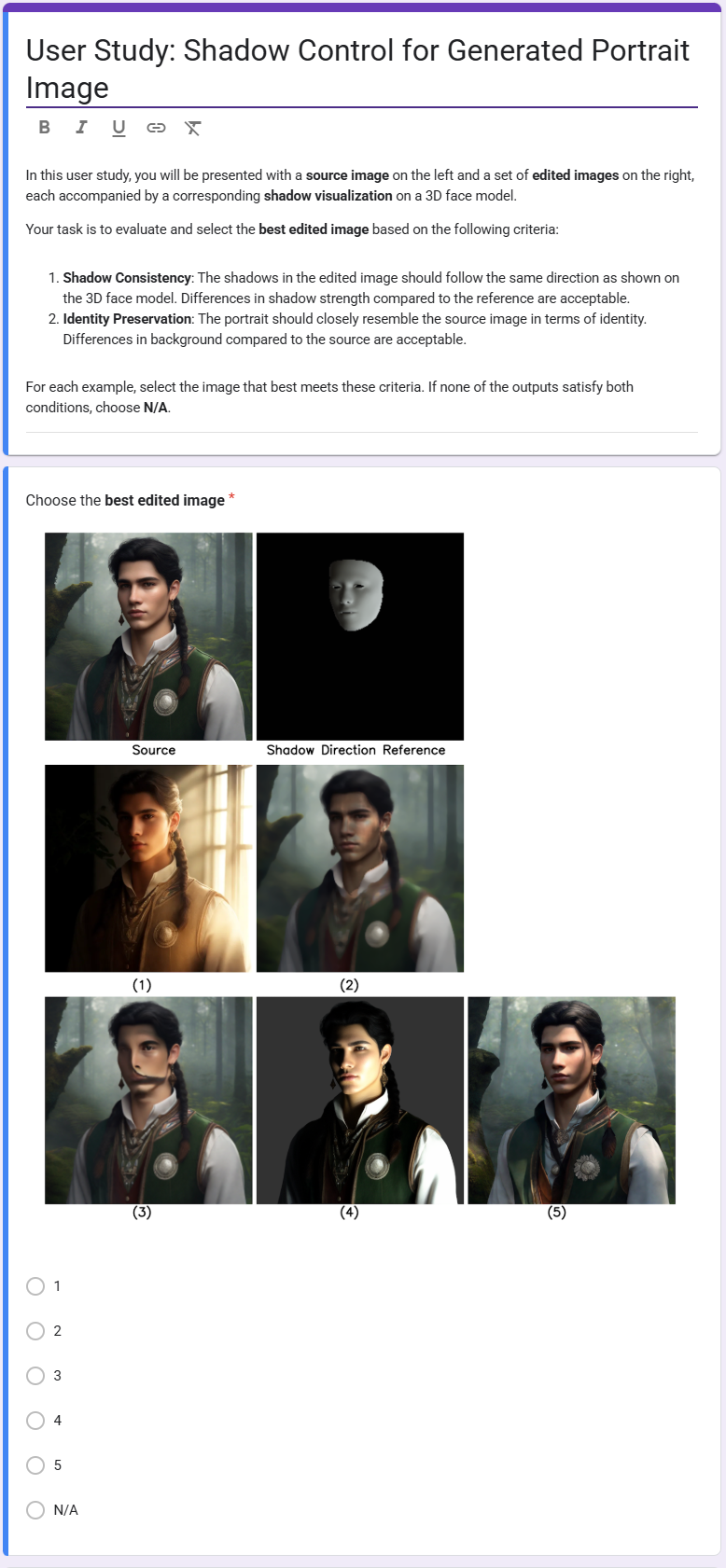}
    \caption{ Example of the user study.}
    \label{supp_fig:user_study}
\end{figure}

\begin{figure*}[tbh!]
    \centering
    \includegraphics[width=\linewidth]{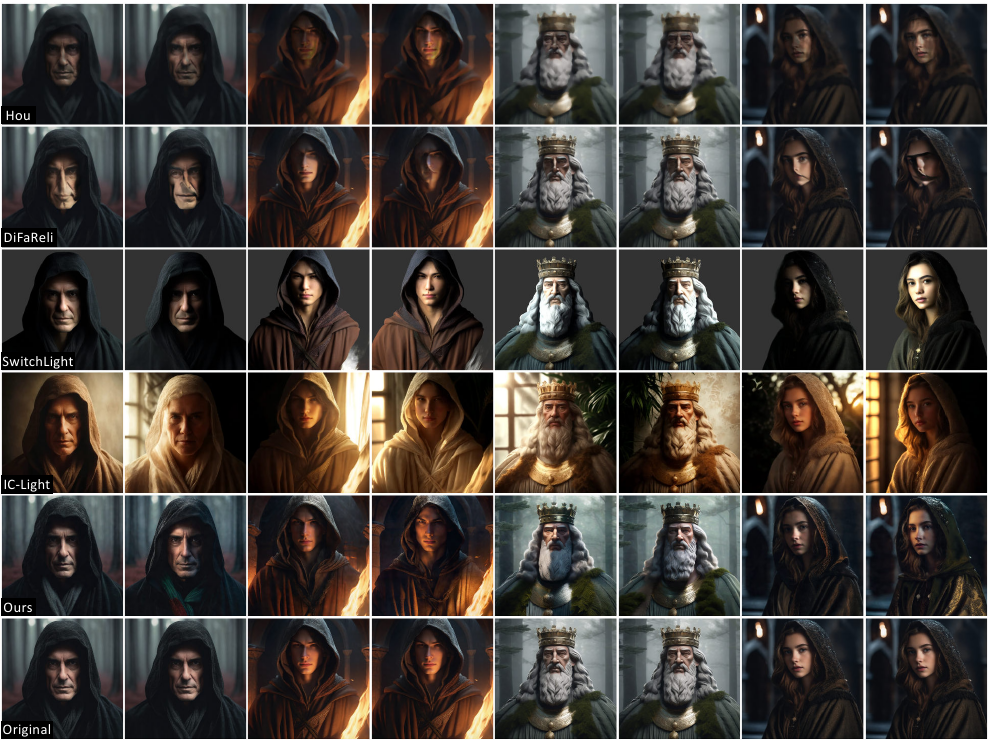}
    \caption{Comparison of portrait relighting across different editing methods on diverse artistic styles. Each row corresponds to a different method, while each column maintains left and right lighting direction for convenient visual comparison. For SwitchLight, we create a related environment map to simulate directional lighting.}
    \label{supp_fig:re_lighting_figure_supp_01}
\end{figure*}

\begin{figure*}[tbh!]
    \centering
    \includegraphics[width=\linewidth]{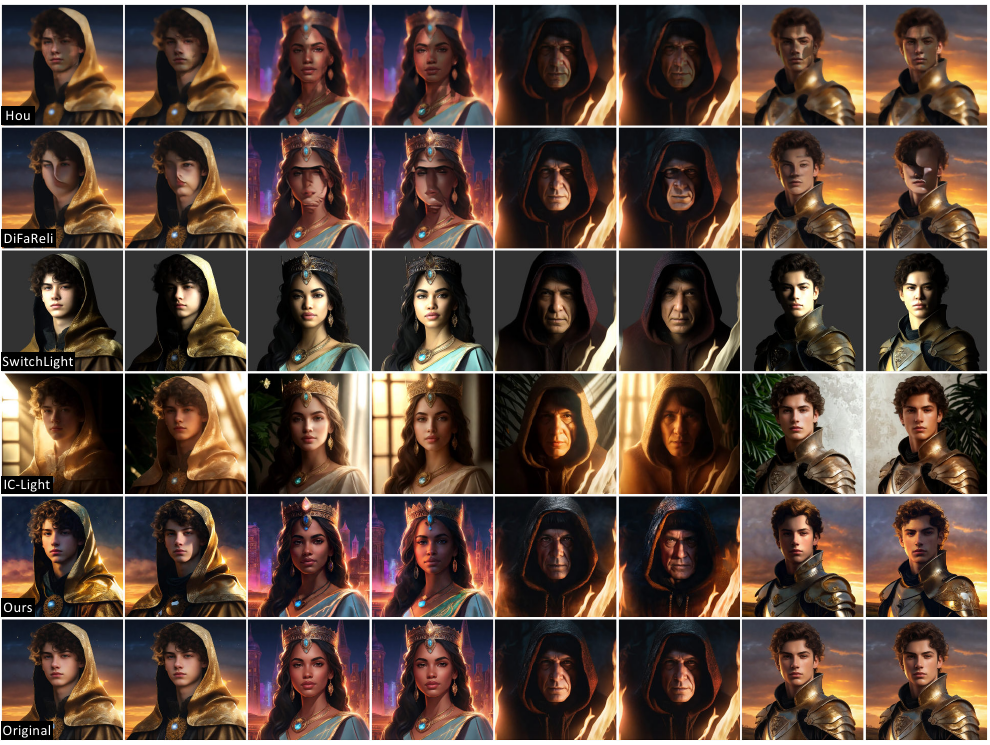}
    \caption{Comparison of portrait relighting across different editing methods on diverse artistic styles. Each row corresponds to a different method, while each column maintains left and right lighting direction for convenient visual comparison. For SwitchLight, we create a related environment map to simulate directional lighting.}
    \label{supp_fig:re_lighting_figure_supp_02}
\end{figure*}

\begin{figure*}[tbh!]
    \centering
    \includegraphics[width=\linewidth]{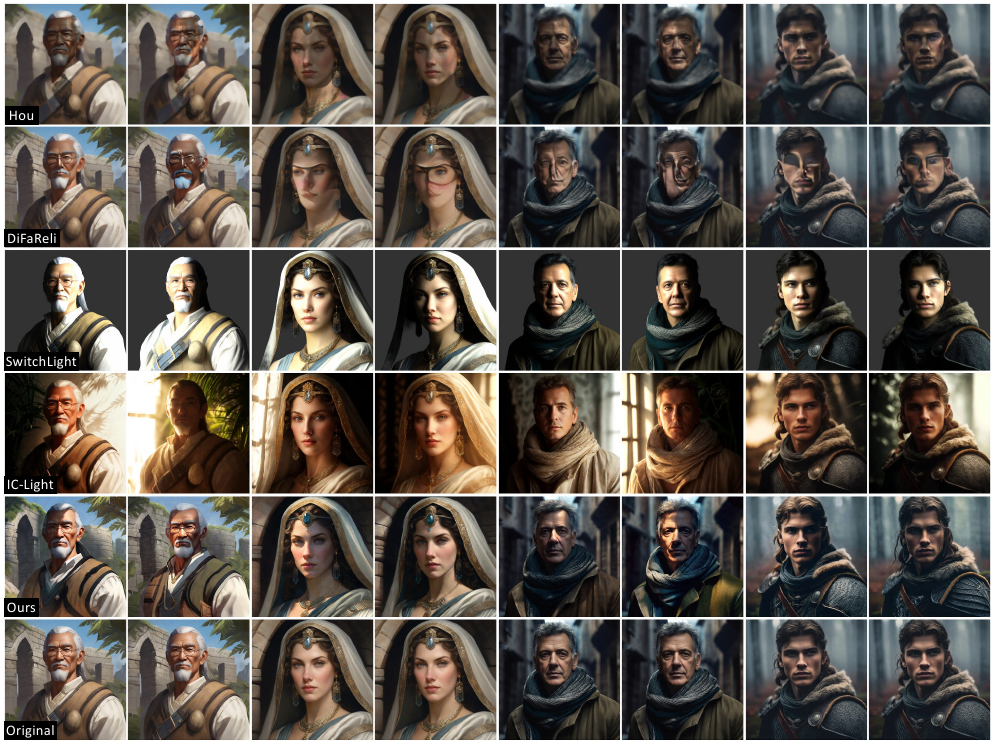}
    \caption{Comparison of portrait relighting across different editing methods on diverse artistic styles. Each row corresponds to a different method, while each column maintains left and right lighting direction for convenient visual comparison. For SwitchLight, we create a related environment map to simulate directional lighting.}
    \label{supp_fig:re_lighting_figure_supp_03}
\end{figure*}

\begin{figure*}[tbh!]
    \centering
    \includegraphics[width=\linewidth]{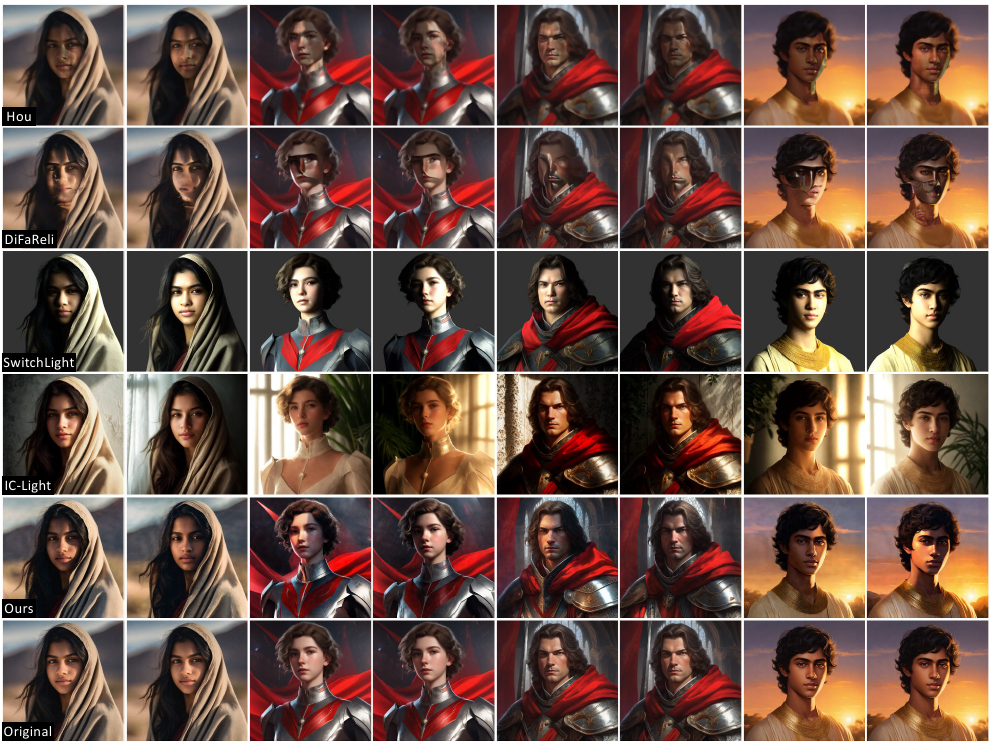}
    \caption{Comparison of portrait relighting across different editing methods on diverse artistic styles. Each row corresponds to a different method, while each column maintains left and right lighting direction for convenient visual comparison. For SwitchLight, we create a related environment map to simulate directional lighting.}
    \label{supp_fig:re_lighting_figure_supp_04}
\end{figure*}

\begin{figure*}[tbh!]
    \centering
    \includegraphics[width=\linewidth]{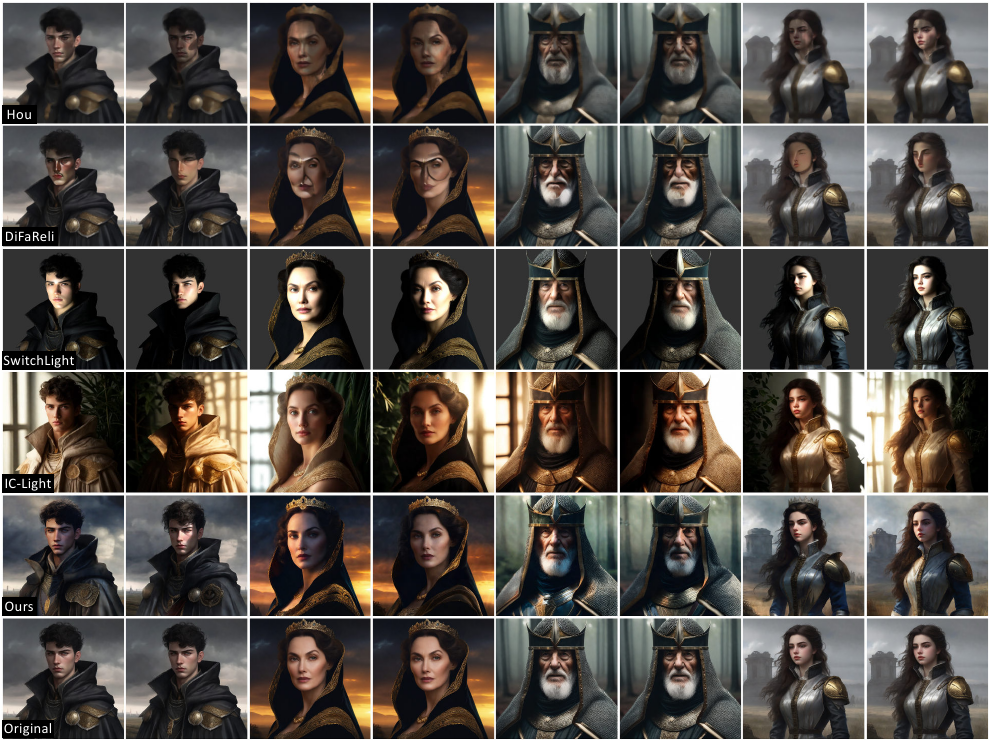}
    \caption{Comparison of portrait relighting across different editing methods on diverse artistic styles. Each row corresponds to a different method, while each column maintains left and right lighting direction for convenient visual comparison. For SwitchLight, we create a related environment map to simulate directional lighting.}
    \label{supp_fig:re_lighting_figure_supp_05}
\end{figure*}

\begin{figure*}[tbh!]
    \centering
    \includegraphics[width=\linewidth]{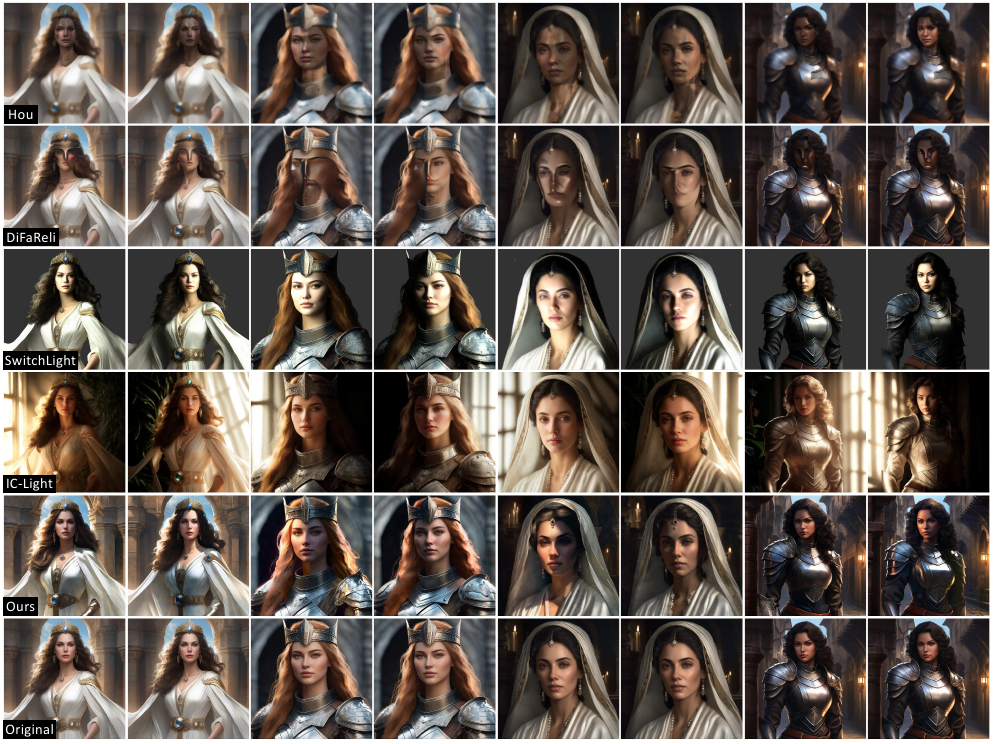}
    \caption{Comparison of portrait relighting across different editing methods on diverse artistic styles. Each row corresponds to a different method, while each column maintains left and right lighting direction for convenient visual comparison. For SwitchLight, we create a related environment map to simulate directional lighting.}
    \label{supp_fig:re_lighting_figure_supp_06}
\end{figure*}

We conduct a user study as shown in Figure \ref{supp_fig:user_study}. We ask the
users to choose the outputs based on specific criteria: content
preservation and text reflection. We show 20 samples and the
outputs of four models: ours. IC-Light. SwitchLight, Hou, and DiFaReLi. The image used in the user study can be seen in Figures \ref{supp_fig:re_lighting_figure_supp_01}–\ref{supp_fig:re_lighting_figure_supp_06}.

\section{Text Prompt Templates and Examples}
\label{supp:text_prompts}

To ensure controlled portrait generation while maintaining consistency in identity and lighting, we designed structured text prompts tailored for our diffusion model. These prompts balance diversity in artistic styles while minimizing interference from excessive accessories, complex clothing, or elaborate backgrounds, which could affect shadow manipulation.

\subsection{Prompt Template}
Our prompts follow a structured format to enforce consistency in composition and lighting conditions. The template is as follows:

\textit{A [STYLE] close-up portrait of a [AGE] [GENDER] with [ETHNICITY] and [FACE\_SHAPE] features. Wearing [SIMPLE\_CLOTHING].}

Where:
\begin{itemize}
    \item \textbf{STYLE}: Specifies the artistic style (e.g., oil painting, cinematic, gothic, fantasy).
    \item \textbf{AGE}: Defines the subject's age category (e.g., young, middle-aged, elderly).
    \item \textbf{GENDER}: Indicates gender identity (e.g., man, woman).
    \item \textbf{ETHNICITY}: Ensures diversity in generated subjects (e.g., Asian, African, Nordic, Mediterranean).
    \item \textbf{FACE\_SHAPE}: Controls facial structure (e.g., angular, round, chiseled).
    \item \textbf{SIMPLE\_CLOTHING}: Limits clothing complexity (e.g., dark tunic, plain robe, leather vest) to preserve identity consistency.
    \item \textbf{BACKGROUND}.
\end{itemize}

\subsection{Example Prompts}
To illustrate the variety of generated portraits, we provide a few example prompts:

\begin{itemize}
    \item \textit{A gothic close-up portrait of a young man with Nordic and chiseled features. Wearing a dark tunic. A blurred studio background.}
    \item \textit{A cinematic close-up portrait of an elderly woman with African and angular features. Wearing a simple robe. A blurred studio background.}
    \item \textit{A fantasy close-up portrait of a middle-aged warrior with Eastern European and strong features. Wearing a leather vest. A blurred studio background.}
    \item \textit{A Renaissance-inspired close-up portrait of a young queen with Mediterranean and delicate features. Wearing an embroidered cloak. A blurred studio background.}
\end{itemize}

These structured prompts allow controlled generation of diverse portraits while ensuring identity preservation, shadow consistency, and style variety.

\begin{figure}[]
    \centering
    \includegraphics[width=\linewidth]{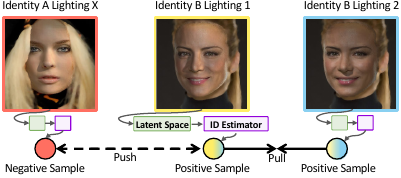}
    \caption{\textbf{Training and Mechanism of the Identity Estimator.}  The Identity Estimator ensures identity consistency during shadow manipulation. In the training phase, we use three images: two with the same identity but different lighting conditions, and a third with a random identity. Identity feature maps (represented by colored spheres) are independently extracted for each image, following a similar process to the Shadow-Depth Estimator. The triplet loss minimizes the distance between features of the same identity (positive samples) and maximizes the distance from features of different identities (negative samples), enabling the Identity Estimator to effectively distinguish identities. During latent optimization in the inference phase, a reference feature is first generated from a text prompt to guide subsequent shadow manipulations for that prompt.}
    \vspace{-10pt}
    \label{fig:id_preservation}
\end{figure}

\end{document}